# Explaining Explaining

Sergei Nirenburg[1], Marjorie McShane[1], Kenneth W. Goodman[2], and Sanjay Oruganti[1]

[1] *Department of Cognitive Science, Rensselaer Polytechnic Institute, 110 8th St. Troy, NY 12180-3590, USA*

[2] *Institute for Bioethics and Health Policy, Miller School of Medicine, University of Miami, P.O. Box 016960 (M-825) Miami, FL 33101, USA*

*{nirens, mcsham2, orugas2}@rpi.edu, KGoodman@med.miami.edu*

Abstract:     Explanation is key to people having confidence in high-stakes AI systems. However, machine-learning-based systems – which account for almost all current AI – can't explain because they are usually black boxes. The *explainable AI* (XAI) movement hedges this problem by redefining "explanation". The *human-centered explainable AI* (HCXAI) movement identifies the explanation-oriented needs of users but can't fulfill them because of its commitment to machine learning. In order to achieve the kinds of explanations needed by real people operating in critical domains, we must rethink how to approach AI. We describe a hybrid approach to developing cognitive agents that uses a knowledge-based infrastructure supplemented by data obtained through machine learning when applicable. These agents will serve as assistants to humans who will bear ultimate responsibility for the decisions and actions of the human-robot team. We illustrate the explanatory potential of such agents using the under-the-hood panels of a demonstration system in which a team of simulated robots collaborates on search task assigned by a human.

## 1   INTRODUCTION

Explanation is clearly one of Marvin Minsky's "suitcase" words "that we use to conceal the complexity of very large ranges of different things whose relationships we don't yet comprehend" (Minsky, 2006, p. 17). The Stanford Encyclopedia of Philosophy (https://plato.stanford.edu/) includes detailed entries on mathematical, metaphysical, and scientific explanation, and a separate one on causal approaches to the latter. Specialist philosophical literature discusses Carl Hempel's (1965) deductive-nomological model of explanation and the rival inductive-statistical approaches. Explanation is also discussed in other disciplines, such as psychology (e.g., Lombrozo, 2010). Special attention is also paid to the differences between explainability, interpretability, transparency, explicitness, and faithfulness (e.g., Rosenfeld & Richardson, 2019). Recent years have also seen a pronounced interest in developing novel theories of explanation (Yang, Folke, & Shafto, 2022; Rizzo et al., 2023).

Explanation in AI has a long history as well. Arguably the first AI-related contribution was Craik (1943). Kenneth Craik was a psychologist and an early cyberneticist whose work influenced AI and cognitive science (Boden, 2006, pp. 210-218). His book, entitled "The Nature of Explanation," discusses a variety of the senses of this suitcase word and, among other things, stresses the distinction between causal explanation in terms of a formal world model (what would later be termed ontology) and statistical explanation, which seeks to explain by pointing out uninterpreted relations among observable entities.

The above distinction still remains in the spotlight today. Most current generative AI systems are black boxes whose functioning cannot be explained in normal human terms. For certain applications, this is not a problem:

1.  Non-critical AI capabilities – such as GPS systems, machine translation systems, and search engines – are widely and happily employed by end users who don't require explanations.
2.  AI capabilities that emulate physical rather than cognitive capabilities – such as robotic movement and speech recognition – are incompatible with the kinds of user-elucidating explanations we address here. That is, we all understand that it would be pointless to ask a robot exactly how it extends its arm or keeps its balance when walking on uneven surfaces.

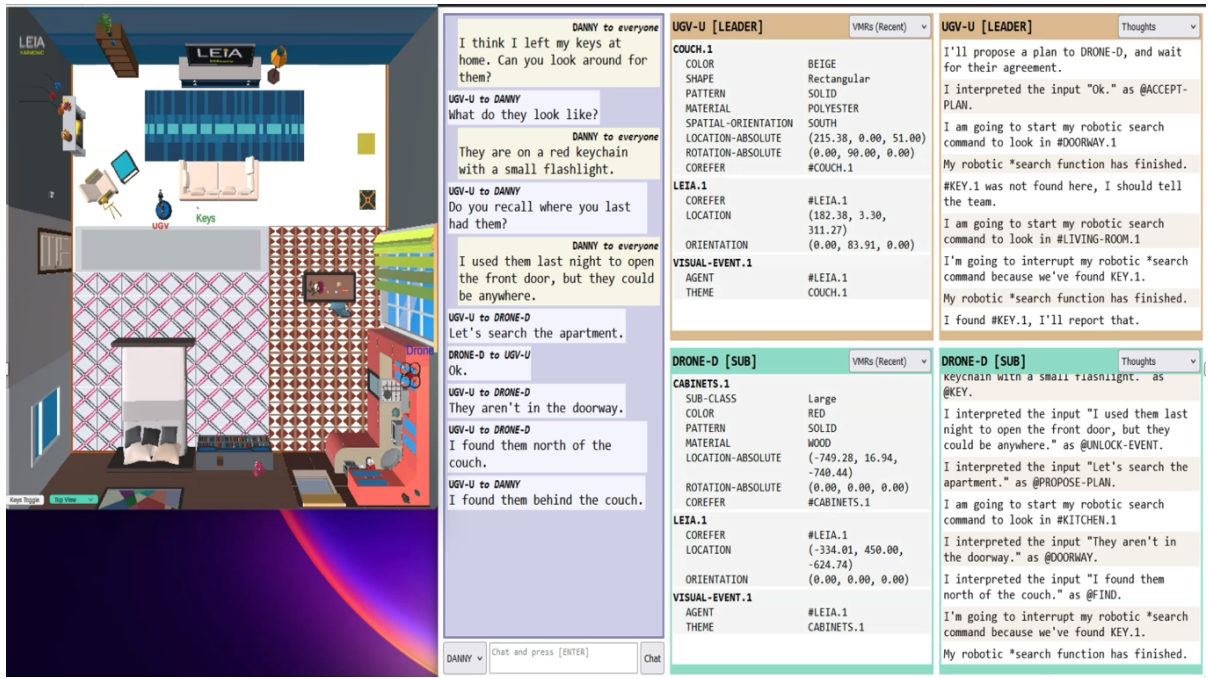

Fig. 1. The demonstration view of the robotic search-and-retrieve system at the end of the system run. The simulation window is to the left, the dialog window is in the middle, and four of many available under-the-hood panels are to the right. This figure is intended only to show the look-and-feel of the system; subsequent figures of individual components will be legible.

3. AI systems that emulate cognitive capabilities (e.g., X-ray analysis systems) can be useful, despite their lack of explainability, as long as they are leveraged as orthotic assistants rather than autonomous systems. This requires that domain experts understand the reliability and deficits of the technology well enough to judiciously incorporate it into their workflow (Chan & Siegel, 2019; Nirenburg, 2017).

By contrast, lack of explainability *is* a problem for ML-based AI systems that are designed to operate autonomously in critical domains. For example, as of June 2021, the FDA cleared 343 AI/ML-based medical devices, with over 80% of clearances occurring after 2018 (Matzkin, 2021). This supply of new AI systems has continued unabated even though their adoption has been less than enthusiastic. Fully 70% of these devices offer radiological diagnostics and typically claim to exceed the precision and efficiency of humans. But, according to Gary Marcus (2022), as of March 2022, "not a single radiologist has been replaced." So, regulators keep approving systems whose operation cannot be explained, and developers keep hoping that their systems, though unexplainable, will be adopted. (For further

discussion, see McShane, Nirenburg, & English, 2024, Section 2.7.1.)

The unexplainability problem has been addressed in earnest by the *Explainable-AI* (XAI) movement (Finzel et al., 2023; Bodria et al., 2021; Cambria et al., 2022; Nagahisarchoghaei et al., 2023; Schwalbe & Finzel, 2024), but the results are, in general, less than satisfying (Barredo et al., 2020). XAI investigators hedge the explainability problem by redefining "explanation" (Gunning, 2017; Mueller et al., 2019). XAI research does not seek to explain how systems arrived at their output. Instead, it concentrates on "*post hoc* algorithmically generated rationales of black-box predictions, which are not necessarily the actual reasons behind those predictions or related causally to them... [and which] are unlikely to contribute to our understanding of [a system's] inner workings" (Babic et al., 2021).

The related *human-centered explainable AI* (HCXAI) movement, for its part, identifies the explanation-oriented needs of users but is hampered in fulfilling them because of its commitment to machine learning (Babic et al., 2021; Ehsan et al., 2022; Liao & Varshney, 2022).

The solution to the problem of unexplainable ML-based AI is not to keep trying to square that circle: generative AI techniques are, and will remain, unexplainable. The solution is to step back and reconsider how to develop AI systems so that they are explainable to the degree, and in the ways, that are necessary for different types of applications.

Intelligent behavior by humans and AI agents involves a variety of different capabilities of perception, reasoning, decision-making, and action. Some of them are arguably better fit to be implemented using generative AI approaches, some others, by symbolic AI approaches. Therefore, hybrid AI systems are better suited for comprehensive (non-silo) applications than either of the above approaches alone. This observation was first made over 60 years ago (Minsky, 1961) and has finally received due attention in the field. Indeed, hybrid "neurosymbolic" AI architectures are at present one of the foci of work in AI (Hitzler et al., 2023).

Our team is developing a family of hybrid cognitive systems that we call LEIAs: Language-Endowed Intelligent Agents. Our work is a part of the movement toward integrating empirical (deep learning-based) and deductive/rational (knowledge-based) approaches to building intelligent agent systems. Explanation is an important component of such systems (de Graaf et al., 2021).

We believe that explanations are needed, first and foremost, for those aspects of agent functioning that end users want to be explained, so that they will come to trust their agent collaborators. (Granted, explanations are also useful for AI developers. But enhancing the efficiency of system development is a different concern from supplying systems with essential explanatory capabilities.) It follows that generative AI methods are most useful for implementing system modules that don't need to be explained, such as motor control and uninterpreted perception. For any process that requires explainable reasoning (perception interpretation, decision-making, action specification, etc.), and for any application where confidence in system output is important, black-box methods, such as large language models (LLM), are not a good fit. Accordingly, LEIAs use LLMs whenever this simplifies or speeds up work on tasks that do not require explanation or for which errors will not undermine the reliability of agent behavior.

In what follows we briefly illustrate some of the explanatory capabilities of LEIAs.

## 2   LEIAs

The LEIA program of R&D is a theoretically grounded, long-term, effort that has two main emphases: developing cognitive robotic systems whose capabilities extend beyond what machine learning alone can offer, and earning people's trust in those systems through explainability (McShane, Nirenburg, & English, 2024).

LEIAs are implemented in a dual-control cognitive-robotic architecture that integrates strategic, cognitive-level decision-making with tactical, skill-level robot control (Oruganti et al., 2024). The strategic (cognitive) layer relies primarily on knowledge-based computational cognitive modeling for interpreting perceptive inputs, reasoning, decision-making, learning, etc. The tactical (robotic, skill-level) module relies on data-driven tools for recognizing perceptive inputs and rendering actions.

LEIAs can explain their operation because they are configured using human-inspired computational cognitive modeling. Their explanations make clear the relative contributions of symbolic and data-driven methods, which is similar to a human doctor explaining a recommended procedure using both causal chains, such as how the procedure works, and population-level statistics, such as the percentage of patients for whom it is curative.

## 3   EXPLANATION VIA UNDER-THE-HOOD PANELS

As detailed in McShane, Nirenburg, and English (2024, chapter 8, "Explaining"), there are many things that a LEIA can explain (what it knows, how it interpreted an input, why it made a given decision, etc.) and there are many ways to present explanations to people. Although the most obvious way is through language, other expressive means can be even more useful in some contexts. One such way is by dynamically showing traces of system operation using what we call under-the-hood panels.

We first introduced under-the-hood panels in the Maryland Virtual Patient (MVP) proof-of-concept clinician training application (McShane et al., 2008; McShane & Nirenburg, 2021). There, the under-the-hood panels showed traces of the physiological simulation of the virtual patient, the patient's interoception, its thoughts, the knowledge it learned, and how it interpreted text inputs from the user, who was playing the role of attending physician. These insights into system functioning were geared toward earning the trust of medical educators, who would ultimately need to choose to incorporate such a system into their pedagogical toolbox.

## 3.1 A search-and-retrieve request

We will illustrate the explanatory power of under-the-hood panels using a new system (Oruganti et al., 2024; Oruganti et al., submitted) in which two simulated robots, a drone and a ground vehicle, work as a team to fulfill a search-and-retrieve request by a person (Fig. 1). A human named Danny, who is located remotely, asks the team of robots – a drone and ground vehicle (UGV) – to find keys that he lost in his apartment. Danny communicates with the UGV, since it is serving as the robotic team leader with the drone as its subordinate. The full dialog, which we'll walk through, is shown in Fig. 2.

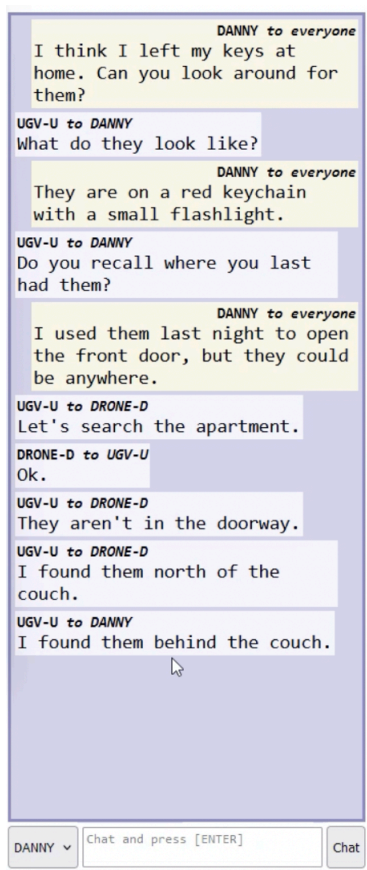

Fig. 2. The full dialog in the demo. The prompt for typing is at the bottom.

When Danny issues his request, both robots semantically interpret the input, resulting in text-meaning representations (TMRs) that are written in the ontologically-grounded metalanguage used for all agent knowledge, memory, and reasoning. This metalanguage uses unambiguous ontological concepts (not words of English) and their instances, described by ontologically-grounded properties and values. Ontological concepts are written in small caps to distinguish them from words of English, and their instances are indicated by numerical suffixes. The process of language understanding is complicated, as detailed in McShane and Nirenburg (2021).

Fig. 3 shows the UGV's TMR of Danny's request and Fig. 4 shows its subsequent thoughts, which are natural language traces of its reasoning for the benefit of humans. (It reasons in the ontological metalanguage.)

| UGV-U [LEADER] | | TMRs (Recent) ⌄ |
| --- | --- | --- |
| KEY.1 | | |
| | CARDINALITY | >,1 |
| | COREFER | TMR.3/KEY.1,#KEY.1 |
| LEIA.1 | | |
| | COREFER | #LEIA.1 |
| REQUEST-ACTION.1 | | |
| | BENEFICIARY | LEIA.1 |
| | THEME | SEARCH-FOR-LOST-OBJECT.1 |
| | AGENT | #HUMAN.1 |
| SEARCH-FOR-LOST-OBJECT.1 | | |
| | AGENT | LEIA.1 |
| | THEME | KEY.1 |
| | TIME | >,FIND-ANCHOR-TIME |

Fig. 3. The UGV's interpretation of Danny's request.

| UGV-U [LEADER] | Thoughts ⌄ |
| --- | --- |
| I interpreted the input "I think I left my keys at home. Can you look around for them?" as @REQUEST-ACTION. | |
| DANNY wants us to @SEARCH-FOR-LOST-OBJECT. | |

Fig. 4. The UGV's thoughts in response to Danny's request.

Because the UGV has received a request for action, and because it knows that has a helper (the drone), it places a COLLABORATIVE-ACTIVITY on its Agenda (Fig. 5). Before it launches the plan for SEARCH-FOR-LOST-OBJECT, it has to check if its preconditions are met (Fig. 5). The first precondition, knowing the object type, is already met (i.e., keys), but the second and third are not: knowing the keys' features and knowing where they were last seen; so the UGV asks about these things in turn. The reasoning associated with this sequence of actions is shown in Fig. 6.

```
UGV-U [LEADER]                    Agenda (Filtered) ⌄

@COLLABORATIVE-ACTIVITY
  [SELECT-PLAN]   i.e., SEARCH-FOR-LOST-OBJECT
    [PRECONDITIONS]
      @REQUEST-OBJECT-TYPE
      @REQUEST-OBJECT-FEATURES
      @REQUEST-LAST-SEEN-AT
      @REQUEST-LOCATION-CONSTRAINED
  @PROPOSE-PLAN
  [RUN-PLAN]
    @SEARCH-FOR-LOST-OBJECT
```

Fig. 5. The UGV's agenda while it is fulfilling
preconditions for searching for the keys.

```
UGV-U [LEADER]                    Thoughts ⌄

DANNY wants us to @SEARCH-FOR-LOST-OBJECT.

We'll use @SEARCH-FOR-LOST-OBJECT to
satisfy the goal.

There are some preconditions for @SEARCH-
FOR-LOST-OBJECT we need to satisfy first.

I need to learn more about #KEY.1's
features.

I interpreted the input "They are on a red
keychain with a small flashlight." as @KEY.

It would be useful to know where #KEY.1 was
last seen.

I interpreted the input "I used them last
night to open the front door, but they
could be anywhere." as @UNLOCK-EVENT.
```

Fig. 6. The UGV's thoughts as it fulfilling preconditions
for searching for the keys.

The UGV then proposes the plan of searching the
apartment to the drone, the drone agrees, and it
launches a plan to do that. Its thoughts – including
those running up to this move – are shown in Fig. 7.

```
DRONE-D [SUB]                     Thoughts ⌄

...                keys at home. Can you look around
them?" as @REQUEST-ACTION.

DANNY wants us to @SEARCH-FOR-LOST-OBJECT.

I'm going to wait for a plan from my team
leader.

I interpreted the input "They are on a red
keychain with a small flashlight." as @KEY.

I interpreted the input "I used them last
night to open the front door, but they
could be anywhere." as @UNLOCK-EVENT.

I interpreted the input "Let's search the
apartment." as @PROPOSE-PLAN.

I am going to start my robotic search
command to look in #KITCHEN.1
```

Figure 7. The drone's thoughts leading up to and including
its starting to search the apartment.

Since the robots in this simulation are decentralized,
each having its own cognitive layer, the drone
independently carries out much of the same reasoning

as the UGV. (Note that our architecture also permits
centralized robots that share a cognitive layer.)

Having agreed upon a plan, the UGV and the
drone leave their charging stations, highlighted in
green in Fig. 8, and begin searching the apartment.

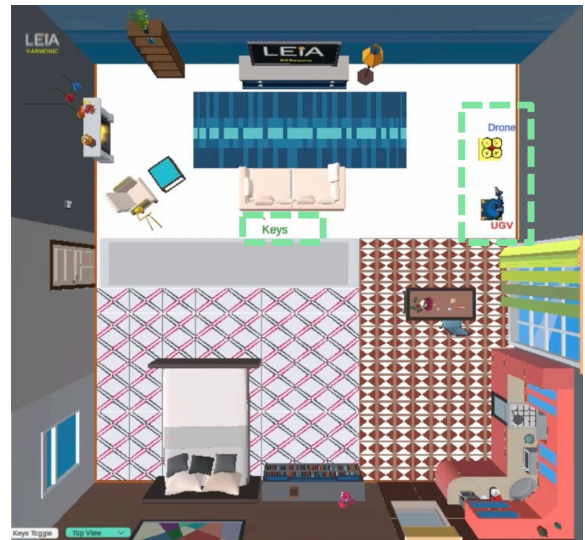

Figure 8. The room layout just before the robots being
searching, with the robots and keys shown in green boxes.

The robots are equipped with sensors to detect,
identify, and tag objects, and to combine this data to
localize objects and themselves. Interpreted traces of
what they are seeing are shown in visual meaning
representations (VMRs) in the associated under-the-
hood panel. VMRs are similar in form and content to
TMRs since, no matter what an agent perceives or
which mode of perception it uses (hearing, vision,
etc.), it has to interpret the stimuli into ontologically-
grounded knowledge that feeds its reasoning. Fig. 9
shows a moment when the UGV is looking at a
particular spot on the blue-green striped carpet.

```
UGV-U [LEADER]                    VMRs (Recent) ⌄

CARPET.1
  SUB-CLASS          Long
  COLOR              BLUE-GREEN
  PATTERN            STRIPES
  MATERIAL           JUTE
  DIMENSIONS         10x2
  LOCATION-ABSOLUTE  (510.00, 0.00, 23.00)
  ROTATION-ABSOLUTE  (0.00, 90.00, 0.00)
  COREFER            #CARPET.1
LEIA.1
  COREFER            #LEIA.1
  LOCATION           (555.75, 3.30, 53.83)
  ORIENTATION        (0.00, 172.36, 0.00)
VISUAL-EVENT.1
  AGENT              #LEIA.1
  THEME              CARPET.1
```

Fig. 9. A visual meaning representation (VMR).

The robots engage in a search strategy involving waypoints, zones and sub-zones that are pre-designated for the apartment environment (Oruganti et al., 2024). The search action is triggered through action commands from the strategic layer but the search itself is controlled by the tactical (robotic) layer. The cognitive (strategic) module knows which zones exist but does not guide how the robots maneuver through those zones. The simulation system is equipped with timing strategies and modules to ensure process and data synchronization between the tactical and strategic layers.

Searching each zone is a subtask of the plan FIND-LOST-OBJECT. After completing each subtask – i.e., searching each zone – each robot reports to the other one about whether it was successful, which is driven by the COLLABORATIVE-ACTIVITY plan.

When the team leader finds the keys, it ceases searching and first reports this to its subordinate and then to Danny. The trace of this reasoning is shown in Fig. 9. It uses different formulations for each of them because its language generation system (whose traces are not shown in this demo system) is designed to mindread its interlocutors and present information in the most useful way for them. Whereas these robots operate in terms of cardinal directions, making *north of the couch* a good descriptor, most humans prefer relative spatial terms like *behind the couch*.

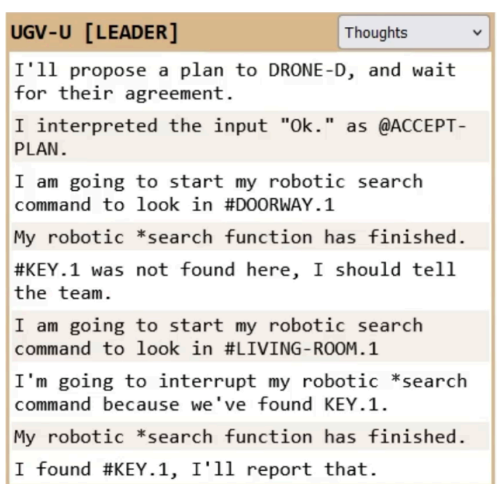

Figure 10. The UGV's thoughts when it finds the keys and decides to report that.

# 4    CONCLUSIONS

Explainability is essential to critical applications, and in order for systems to be truly explanatory, they must first of all understand what they are doing. This requires that they be grounded in high-quality knowledge bases that optimally integrate causal and correlational reasoning.

This paper focused on explanation via traces of system operation using under-the-hood panels. The panels selected for this demo displayed the agents' interpretation of language inputs and visual stimuli, their reasoning, and their agenda. Much more could be shown if target users found that helpful: the agents' ontologies, episodic memories, lexicons, decision-making about language generation, and so on. The under-the-hood panels do not attempt to capture unexplainables that are implemented using machine learning, such as what drives robotic movement or the robots' approach to searching a space.

In the current benchmark-driven climate, under-the-hood panels offer an alternative standard of system evaluation.

Under-the-hood panels are just one mode of explanation for LEIAs. The other primary one is language. The many things that a LEIA can explain using language are detailed in Chapter 8 of McShane, Nirenburg, and English (2024).

Although the theoretical, methodological, and knowledge prerequisites for explanation by LEIAs are quite mature, this doesn't mean that all problems associated with explanation are solved.

Consider the example of physicians explaining relevant aspects of clinical medicine to patients, a capability that was relevant for the MVP clinician-training system mentioned above. The task has two parts: deciding *what* to say and *how* to say it. Both of these depend not only medical and clinical knowledge, but also on the salient features of patients, such as their health literacy (as hypothesized by the physician), their interest in medical details, their ability to process information based on their physical, mental, and emotional states, and so on. Identifying these salient features involves *mindreading* (Spaulding, 2020) – also known as *mental model ascription*. For example, an explanation may be presented in many different ways:

- as a causal chain: "You feel tired because of an iron deficiency."
- as a counterfactual argument: "If you hadn't stopped taking your iron supplement you wouldn't be feeling so tired."
- as an analogy: "Most people find it easier to remember to take their medicine first thing in the morning; you should try that."
- using a future-oriented mode of explanation: "If you take your iron supplement regularly, you should feel much more energetic."

Moreover, explanations are not limited to speech – they can include images, videos, body language, live demonstration, and more. Overall, generating explanations tailored to particular humans is a difficult task. However, as with all other aspects of cognitive modeling, simplified solutions hold promise to be useful, particularly given the well-established fact that adding more content to an explanation does not necessarily make it better (cf. the discussion of decision-making heuristics in Kahneman, 2011).

## ACKNOWLEDGEMENTS

This research was supported in part by Grant #N00014-23-1-2060 from the U.S. Office of Naval Research. Any opinions or findings expressed in this material are those of the authors and do not necessarily reflect the views of the Office of Naval Research.